\documentclass{article} 
\usepackage{formatting/iclr2027_conference,times}

\usepackage{formatting/packages}
\usepackage{formatting/hadi-colors}
\usepackage{formatting/hadi-macros-2025}
\usepackage{formatting/aliases}
\usepackage{comment}

\title{Conjugate Compilation: Self-Improving Agentic Co-Compilation for Speculative Decoding}

\author{\textbf{Annabelle Sujun Tang$^{1}$, Rohan Mahapatra$^{1}$, Christopher Priebe$^{1}$, Hanyang Xu$^{1}$,}\\
\textbf{Manya Ghobadi$^{2,3}$, Christopher Rossbach$^{4,5}$, Hadi Esmaeilzadeh$^{1}$}\\[0.5ex]
$^{1}$University of California, San Diego \\
$^{2}$Massachusetts Institute of Technology \quad
$^{3}$Systalyze \\
$^{4}$University of Texas at Austin \quad
$^{5}$Microsoft \\[0.5ex]
\texttt{\{sujun,rohan,cpriebe,hanyang\}@ucsd.edu},\\
\texttt{ghobadi@csail.mit.edu, rossbach@cs.utexas.edu, hadi@ucsd.edu}
}

\iclrfinalcopy 

\begin{document}

\maketitle

\begin{abstract}
As LLM adoption surges, scaling model serving requires not only adding more GPUs but also unlocking the substantial capacity stranded in those already deployed.
Speculative decoding brings two models into the serving process: a small draft model proposes tokens that a larger target model verifies.
Yet compiling these models individually leaves their shared execution outside the compiler's optimization boundary.
We therefore ask whether co-locating the draft and target models on the same GPU and co-compiling their execution can reclaim this stranded capacity.
We introduce \Conjugate{}, which reframes speculative decoding as a cross-model compilation problem and jointly optimizes the draft and target models within their shared execution rather than treating each in isolation.
One of our key insights is that cross-model compilation should be informed by the measured performance of the co-located shared execution it produces: an optimization that improves a kernel in isolation need not improve the models when they execute together.
To expand cross-model opportunities, we devise operand layouts that interleave blocks from both models within common tiles, co-locating GEMM and non-GEMM work while preserving their mathematics.
We realize \Conjugate{} through a self-improving agentic search system in which an LLM edits cross-model optimization plans using the compiler and GPU execution as its environment.
Each proposed plan is compiled and evaluated through its co-located shared execution, whose measured score, together with kernel-level verdicts and compiler feedback, informs subsequent proposals, while Monte Carlo tree search accumulates rewards to guide exploration.
Across six single-GPU speculative-decoding configurations on NVIDIA B300 and RTX 5090 GPUs, \Conjugate{} achieves a geometric-mean per-round speedup of $1.37\times$ over conventional execution.
Interleaved kernels raise this speedup to $1.45\times$, reaching $1.59\times$ in the best case.
\Conjugate{} also improves GPU utilization on average by $1.46\times$ on RTX 5090 and $1.50\times$ on B300.
These results demonstrate the potential of cross-model conjugate compilation guided by measurements of the co-located shared execution.

\end{abstract}

\section{Introduction}
\label{sec:introduction}
\begin{wrapfigure}{r}{0.51\textwidth}
  \centering
  \includegraphics[width=0.84\linewidth]{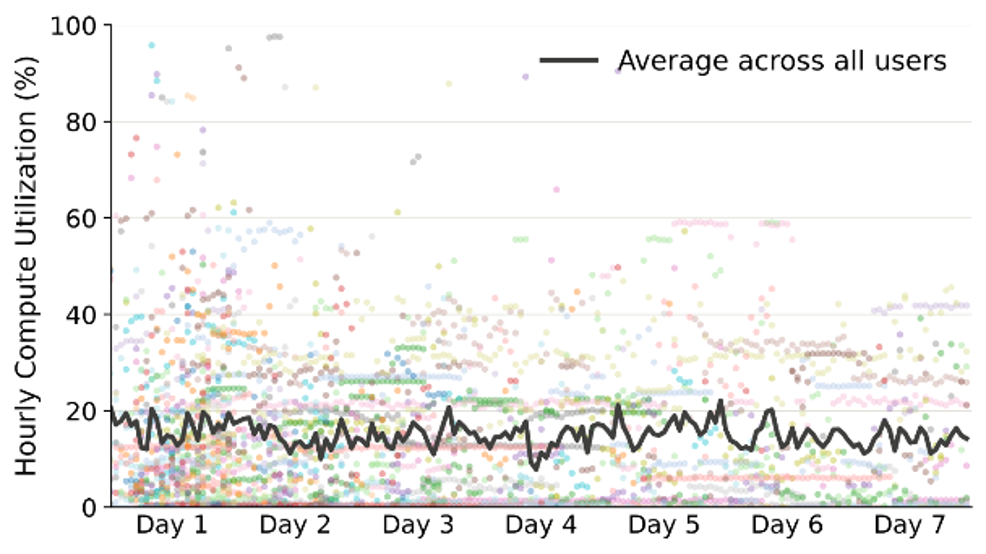}
  \caption{Hourly GPU compute utilization across enterprise users measured with Systalyze Utilyze from May to August 2026. The black line shows the hourly average across users; average utilization over the full dataset is 17.6\%.}
\label{fig:motivation}
\end{wrapfigure}

\vspace{-8pt}
As demand for LLM serving grows, expanding GPU infrastructure is only one way to increase serving capacity~\citep{iea:energy-and-ai:2025}.
Another is to recover the substantial capacity that remains unused within existing deployments.
\fref{fig:motivation} shows one week of hourly GPU compute utilization measurements across enterprise users, illustrating how low average utilization persists despite variation across users and time.
Across 6,000 measurements from 350 users collected between May and August 2026, average GPU compute utilization is 17.6\%, with 94\% of users averaging below 50\%~\citep{utilyze}.
These observations motivate us to examine how execution-informed compilation can turn more of the GPUs already deployed into useful capacity for model serving.

Turning this capacity into useful work depends in part on compilation, which determines how a model's operations and data are organized for GPU execution~\citep{dnn-compiler-survey:tpds:2020}.
The common practice is to compile and optimize each model individually~\citep{conco:ics:2025}.
Yet a single model's operations may provide too little parallel work to fill the GPU or leave compute resources underused while waiting for data~\citep{ios:mlsys:2021,pope:mlsys:2023}.
This limitation motivates us to expand the compiler's optimization boundary: rather than optimize each model in isolation, we bring multiple models into concurrent execution on the same GPU and jointly optimize them informed by their co-located shared execution.

Speculative decoding provides a setting for exploring this opportunity because two models already cooperate to generate each response.
A small draft model proposes several tokens, and a larger target model verifies them in a single forward pass, reducing the number of target-model passes needed for generation~\citep{spec-decode:icml:2023}.
This cooperation, however, does not by itself coordinate how their operations share GPU resources, nor does it bring their co-located shared execution within the scope of compiler optimization~\citep{pearl:iclr:2025,swiftspec:asplos:2026}.
We therefore pose a central research question: without changing either model's underlying computations, can we coordinate and co-compile the draft and target models to reclaim the capacity left unused when they are optimized individually?

Speculative decoding provides a concrete setting for exploring this opportunity because two models already cooperate to generate each response.
A small draft model proposes several tokens, and a larger target model verifies them in a single forward pass, reducing the number of target-model passes needed for generation~\citep{spec-decode:icml:2023}.
Prior approaches primarily exploit this structure as a scheduling, placement, or resource-allocation opportunity, either distributing draft and target execution across multiple GPUs or within a single GPU, while leaving their shared execution outside the compiler's optimization boundary~\citep{pearl:iclr:2025,swiftspec:asplos:2026,faser:arxiv:2026}.
We instead pose a central research question: what becomes possible when we co-locate them on the same GPU and let their shared execution directly inform cross-model compilation.

We introduce \Conjugate{}, which reframes speculative decoding as a cross-model compilation problem by bringing the draft and target models into a common optimization boundary on the same GPU.
Their co-located shared execution uses capacity left idle by either model, but also changes the conditions under which their operations execute.
An optimization that improves a kernel in isolation may increase contention, disrupt beneficial overlap, or otherwise fail to improve the models when they execute together~\citep{gpu-interference:socc:2025}.
One of our key insights is therefore that cross-model compilation should be informed by the measured performance of the co-located shared execution it produces.
\Conjugate{} uses this joint execution feedback to jointly optimize the draft and target models rather than optimizing each model independently.

Optimizing within this co-located shared execution requires a search that can respond to how the draft and target models actually behave together.
We therefore contribute a self-improving agentic system that couples an LLM with Monte Carlo tree search (MCTS)~\citep{reasoning-compiler:neurips:2025}, using the compiler and GPU execution as its environment.
The LLM proposes edits to cross-model compilation plans, which the compiler checks, builds, and tunes before we measure the performance of the resulting co-located shared execution.
We use MCTS as the agent's search medium, organizing explored plans and propagating their measured rewards to guide further exploration, while returning the complete execution score, kernel-level verdicts, and compiler feedback to the LLM to inform subsequent proposals.
This coupling allows the agent to improve its proposals from accumulated execution evidence and directs exploration toward plans that improve the models together.

To give the agentic system choices beyond coordinating separate kernels, we contribute an operand memory layout transformation that brings work from both models into a common kernel.
By interleaving operand blocks within common tiles, we co-locate their work while preserving which operands participate in each computation and where its results belong.
The interleaved kernels support GEMM and non-GEMM operation groups, including RMSNorm~\citep{rmsnorm:neurips:2019} and residual addition~\citep{resnet:cvpr:2016}, without changing their mathematics.
The agent explores which operation groups to interleave and how much work to draw from each model, while the compiler optimizes these transformations as part of the broader cross-model compilation plan.
Because an interleaved kernel that improves in isolation need not improve the models together, \Conjugate{} evaluates these choices using the measured performance of the complete co-located shared execution.

We also contribute an operand memory layout transformation that brings work from both models into a common kernel.
By interleaving operand blocks within common tiles, we co-locate their work while preserving which operands participate in each computation and where its results belong.
The interleaved kernels support GEMM and non-GEMM operation groups, including RMSNorm~\citep{rmsnorm:neurips:2019} and residual addition~\citep{resnet:cvpr:2016}, without changing their mathematics.
The agent explores which operation groups to interleave and how much work to draw from each model, while \Conjugate{} optimizes these choices using the measured performance of their co-located shared execution.

We evaluate \Conjugate{} across six single-GPU speculative-decoding configurations on NVIDIA RTX 5090 and B300 GPUs.
Compared with compiling and optimizing each model in isolation, \Conjugate{} achieves a geometric-mean per-round speedup of $1.37\times$.
Adding selectively interleaved kernels raises this speedup to $1.45\times$, reaching $1.59\times$ in the best case.
\Conjugate{} also improves GPU utilization by $1.46\times$ on RTX 5090 and $1.50\times$ on B300, on average, respectively.
Together, these results show that cross-model conjugate compilation, informed by the measured performance of co-located shared execution, can recover stranded capacity by jointly optimizing multiple models on a single GPU.

\section{\Conjugate: Co-Optimizing Shared Execution}
\label{sec:co-compilation}
\vspace{-8pt}

\begin{figure}[!t]
  \centering
  \includegraphics[scale=0.7]{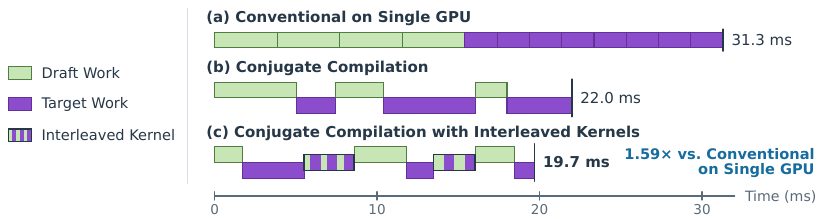}
  \caption{Illustration of (a) conventional execution of speculative decoding, (b) with \Conjugate, and (c) with \Conjugate \xspace    plus Interleaved Kernels. Llama-3.2-1B (draft, $\gamma=4$) and Llama-3.1-8B (target).}
  \label{fig:coscheduled-round}
  \vspace{-0.4cm}
\end{figure}

\subsection{Bringing Drafting and Verification into Shared Execution}
\label{sec:shared-execution}
\frefsub{fig:coscheduled-round}{a} shows conventional execution of speculative decoding on a single GPU.
The green blocks denote draft-model segments and the purple blocks denote target-model segments, with horizontal position representing execution time.
The $\gamma$ draft steps complete before target verification begins, so the green and purple work does not overlap.
If one draft step takes $T_D$ and target verification takes $T_V$, their execution takes
\begin{equation}
    T_{\mathrm{conventional}} = \gamma T_D + T_V.
\end{equation}
This ordering reflects a true dependency within each batch: target verification cannot begin until that batch's draft tokens are available.
The dependency, however, does not extend across batches.
While the target verifies the proposals for batch $b$, the draft model can generate proposals for batch $b+1$.

We use a dependency-preserving multi-stream co-scheduler to expose this concurrency on the same GPU.
\frefsub{fig:coscheduled-round}{b} shows the resulting co-scheduled execution.
Where purple target and green draft blocks overlap along the time axis, segments from the two models execute concurrently; gaps indicate intervals in which only one stream is executing.
We divide both models into ordered segments, ranging from layers and attention/MLP halves down to individual projection stages, and submit the target and draft segments to separate CUDA streams.
Segments within each stream execute in order, while segments on different streams may execute concurrently unless a cross-stream dependency orders them.
Our co-scheduler inserts CUDA events~\citep{cuda-streams} only where such coordination is required, preserving the dependencies within each model while exposing legal overlap.

Formally, let $G_\sigma=(V,E_\sigma)$ denote the co-scheduled round, where $V$ contains the draft and target segments and $E_\sigma$ contains their intra-model dependencies together with any cross-stream edges $\sigma$.
Execution preserves every edge:
\begin{equation}
    t_{\mathrm{start}}(v)
    \ge
    t_{\mathrm{finish}}(u),
    \qquad
    \forall (u,v)\in E_\sigma .
    \label{eq:dependency-constraint}
\end{equation}
For co-scheduling alone, $\sigma=\emptyset$; cross-model compilation decisions may later introduce cross-stream edges when their placement requires them.
We capture both streams and their dependencies in a single CUDA graph and replay the complete co-scheduled round as one unit, removing host launch overhead while preserving the dependency structure.
This co-scheduled round realizes the co-located shared execution that \Conjugate~optimizes.
Its measured behavior provides the execution feedback for the cross-model compilation decisions introduced next.

\subsection{Cross-Model \Conjugate}
\label{sec:co-compilation}
\vspace{-4pt}

Once the draft and target models are co-located, compiler decisions interact through their shared execution.
\Conjugate~therefore jointly optimizes cross-model compilation plans and judges them in the execution they change.
This optimization applies even when the two models retain separate kernels on separate CUDA streams.
Their kernel configurations can still be selected according to their effect on the complete shared execution rather than their isolated latency.
We represent these choices through a \emph{joint optimization plan} $\theta$, which captures the cross-model compiler decisions realized within the shared execution.

Realizing a joint optimization plan involves four components:
the \emph{conjugated operation groups} selected for joint execution,
the \emph{interleaving proportions} that determine how much work each operation contributes to this joint execution,
the \emph{kernel configurations} used to execute the resulting work,
and the \emph{scheduling arrangement} that places it into the co-scheduled execution.

\niparagraph{Cross-model conjugated operation groups.}
The agent represents the cross-model transformations in a plan as
\begin{equation}
    \mathcal{G}(\theta)=\{g_1,\ldots,g_n\},
\end{equation}
where each conjugated operation group
\begin{equation}
    g =
    \left(
        f,\{o_k\}_{k=1}^{K_g}
    \right),
    \qquad
    2 \le K_g \le K_{\max},
\end{equation}
contains $K_g$ independent operations from the draft and target models that share a kernel family $f$.
These families include general matrix multiplication (GEMM) as well as row operations such as RMSNorm, residual addition, and softmax.
The grouped operations may differ in role and shape, while operations outside the selected groups retain separate kernels.

\niparagraph{Interleaving proportions.}
In our implementation, a selected group can be realized as a cross-model \emph{interleaved kernel} that executes work from multiple operations in one kernel launch.
The striped blocks in \frefsub{fig:coscheduled-round}{c} show where selected draft and target operations are replaced by such kernels, while the remaining purple and green blocks continue to execute as separate kernels.
For GEMM, each member is decomposed into output-tile and reduction-slice units, and the fraction of work assigned to member $k$ follows its share of the group's arithmetic work:
\begin{equation}
    \pi_k =
    \frac{F_k}
         {\sum_{r=1}^{K_g} F_r},
\end{equation}
where $F_k$ is the arithmetic work of operation $o_k$.
The interleaving proportion is therefore derived from the selected group rather than searched independently.
A descriptor table identifies each unit's member, tile coordinates, reduction slice, and operand addresses, allowing operations with unrelated shapes and separately stored operands to execute together.
Mixing their work within one launch can fill partial waves and other execution capacity that the individual operations would leave underutilized.
Each unit nevertheless reads and accumulates only its own operation's data, preserving every member's computation.
The same principle extends to non-GEMM operations such as RMSNorm and residual addition.
Interleaving is therefore one cross-model transformation available to \Conjugate~rather than the definition of the joint optimization plan itself.

\niparagraph{Kernel configurations.}
For both separate and conjugated kernels, \Conjugate~exposes multiple backend
kernel configurations through TVM.
Rather than selecting configurations solely by isolated kernel latency,
\Conjugate~evaluates candidate configurations within the co-located shared
execution and uses their effect on complete execution performance to guide
selection.
For interleaved GEMMs, TVM explores the same low-level schedule space available
to the corresponding separate kernels, while row-operation kernels are selected
from the schedules exposed by their templates.
Kernel configuration is therefore part of the execution-informed optimization:
even when the draft and target remain separate kernels, their implementations
can be chosen according to how they perform together.

\niparagraph{Scheduling arrangement.}
For a given co-scheduled execution, the execution scheme in \sref{sec:shared-execution} fixes the current segmentation, stream assignment, and dependencies that determine where draft and target work may legally overlap.
The conjugated operation groups determine which target segment $i$ and draft segment $j$ must rendezvous for joint execution.
These pairings induce a joint map $i\mapsto j$ that must be injective and strictly increasing along both execution chains, preserving their relative order and preventing cycles.
A joint node $J(i,j)$ is the scheduling container for this co-compiled work.
Selected groups may be realized by interleaved kernels, shown as striped blocks in \frefsub{fig:coscheduled-round}{c}, while other operations retain separate kernels whose execution remains part of the same shared optimization context.
The target stream joins the corresponding draft work before entering $J(i,j)$, and completion of the joint node releases subsequent draft work through a fork.

\subsection{Judging a Joint Optimization Plan in the Execution It Changes}
\vspace{-4pt}

Let $\Theta_{\mathrm{legal}}$ denote the statically legal joint optimization plans defined in \sref{sec:co-compilation}.
Within this shared execution, \Conjugate \xspace seeks
\begin{equation}
    \theta^\star
    \in
    \arg\min_{\theta\in\Theta_{\mathrm{legal}}}
    T_{\mathrm{round}}(\theta),
\end{equation}
where $T_{\mathrm{round}}(\theta)$ is the measured latency of the complete shared execution produced by joint optimization plan $\theta$.
After the compiler realizes a plan, we additionally verify its numerical outputs against the reference execution before considering its performance.

Crucially, this objective does not decompose into isolated kernel improvements.
Let $\tau(o)$ denote the isolated latency of operation $o$, and $\tau(g)$ the latency of the cross-model implementation of conjugated group $g$.
In general,
\vspace{-0.3cm}
\begin{equation}
    T_{\mathrm{round}}(\theta)
    \neq
    T_{\mathrm{round}}(\emptyset)
    -
    \sum_{g\in\mathcal{G}(\theta)}
    \left(
        \sum_{o_k\in g}\tau(o_k)
        -
        \tau(g)
    \right).
\end{equation}
Shared execution may already hide part of an operation's latency, while a decision in the joint optimization plan can change resource contention, overlap, and synchronization elsewhere in the round.
A faster conjugated group therefore does not necessarily produce a faster shared execution.

We consequently judge each joint optimization plan by the complete shared execution it produces, including the synchronization and transformation overhead introduced by cross-model compilation.
Because the decisions in a joint optimization plan interact, their benefit must be learned from measured shared execution rather than from isolated kernel improvements.
\sref{sec:agentic-search} describes how our self-improving agentic system searches this execution-dependent space using execution feedback.
\section{Self-Improving Agentic Search for Conjugate Compilation}
\label{sec:agentic-search}
\vspace{-6pt}

\begin{figure}[!t]
  \centering
  \includegraphics[scale=0.6]{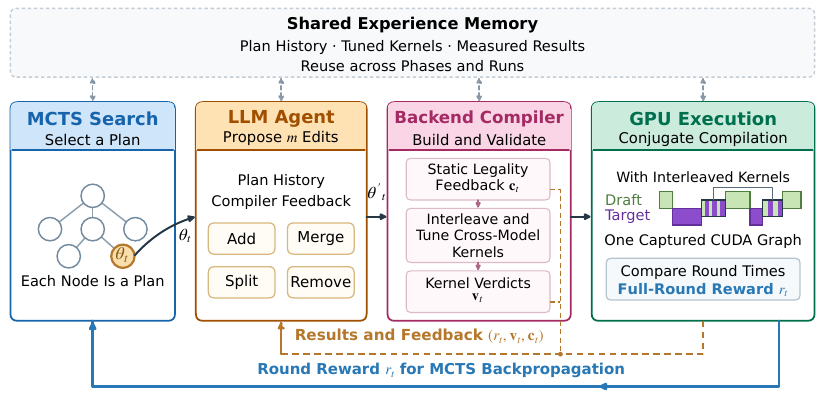}
  \vspace{-0.3cm}
  \caption{Conjugate compilation as a self-improving agentic loop.
}
  \label{fig:agentic-loop}
  \vspace{-0.3cm}
\end{figure}

The joint optimization space is combinatorial and expensive to explore: the value of a plan is known only after compilation, and the GPU measures the resulting shared execution.
In addition, each evaluation produces rich information---including compiler legality feedback, kernel-level verdicts, and complete shared-execution performance---that can guide subsequent optimization.
We therefore build the self-improving agentic loop in \fref{fig:agentic-loop}.
MCTS selects a previously explored joint optimization plan to refine, the LLM agent proposes structured refinements to that plan, and the compiler and GPU evaluate the resulting candidates.
The measured round reward updates the search tree, while compiler feedback and kernel-level verdicts inform subsequent LLM proposals.

\subsection{MCTS-Guided Plan Selection}
\label{sec:mcts}
\vspace{-4pt}

We organize explored joint optimization plans as a Monte Carlo tree search (MCTS), shown on the left of \fref{fig:agentic-loop}.
Each node represents a complete joint optimization plan, with the empty plan $\emptyset$, corresponding to the co-scheduling-only baseline, as the root, and each edge represents one refinement proposed by the LLM agent.
MCTS determines which explored plan should be refined next, balancing further exploration of promising plans against revisiting less-explored alternatives.

Selection follows the upper confidence bound for trees,
\begin{equation}
    \theta_t
    =
    \arg\max_{\theta' \in \operatorname{ch}(\theta)}
    \left[
        \overline{Q}(\theta')
        +
        \beta
        \sqrt{
            \frac{
                \ln\!\left(N(\theta)+1\right)
            }{
                N(\theta')
            }
        }
    \right],
    \label{eq:uct}
\end{equation}
where $\operatorname{ch}(\theta)$ denotes the children of $\theta$, $N(\cdot)$ counts visits, and $\overline{Q}$ is the normalized mean backed-up reward.
Progressive widening limits how quickly new alternatives are introduced, allowing a node to gain a child only while
\begin{equation}
    |\operatorname{ch}(\theta)|
    <
    \left\lceil
        \kappa\left(N(\theta)+1\right)^\alpha
    \right\rceil.
    \label{eq:progressive-widening}
\end{equation}
\vspace{-0.5cm}

Once MCTS selects $\theta_t$, the LLM proposes $m$ candidate refinements (\sref{sec:agent-actions}), which become children of $\theta_t$ as progressive widening permits.
Because each node is an executable joint optimization plan, we evaluate a child directly in the complete shared execution rather than through simulated rollouts and propagate its measured reward through the tree.
Candidates that fail compilation or numerical validation receive the lowest reward, while their failure information remains available to subsequent proposals.

\subsection{LLM-Guided Joint Optimization Proposals}
\label{sec:agent-actions}
\vspace{-4pt}

Given the plan $\theta_t$ selected by MCTS, the LLM agent proposes new joint optimization plans by refining its conjugated operation groups.
The LLM does not generate GPU kernels directly; instead, it operates on the joint optimization plan abstraction defined in \sref{sec:co-compilation}.
The agent has four actions:
\emph{add} creates a new conjugated operation group,
\emph{merge} combines existing groups,
\emph{split} separates members of a group, and
\emph{remove} dissolves a group.
Each action identifies operations and ranges of target layers and draft steps, allowing one proposal to apply the same transformation across multiple locations in the shared execution.

The agent returns $m$ distinct proposals.
Edit types, operation identifiers, ranges, and legal placements are drawn from enumerated choices supplied by the compiler, making each proposal either directly interpretable or rejectable with a specific feedback code.
The LLM proposes these refinements from accumulated optimization experience rather than from the selected plan alone.
Its prompt includes the selected plan and its recent ancestors, their measured round performance, accumulated kernel-level verdicts, compiler feedback, legal placements, and measurements of representative draft--target segment pairs executed separately, concurrently, and conventionally.
By comparing a plan with its ancestors and the feedback they received, the agent can associate changes in measured shared execution with the decisions that produced them and use that evidence to guide subsequent proposals.
The LLM and MCTS thus serve complementary roles: the LLM converts compiler and execution evidence into informed edits, while MCTS determines which explored plans should be refined and which alternatives should be revisited.

\subsection{Learning from Compiler and GPU Feedback}
\label{sec:agent-feedback}
\vspace{-4pt}

The compiler and GPU form the agent's environment.
Evaluating a candidate plan $\theta'_t$ returns
\begin{equation}
    \mathcal{E}(\theta'_t)
    =
    \left(
        r_t,\,
        \mathbf{v}_t,\,
        \mathbf{c}_t
    \right),
    \label{eq:agent-environment}
\end{equation}
where $r_t$ is the measured reward, $\mathbf{v}_t$ contains kernel verdicts, and $\mathbf{c}_t$ contains compiler feedback.

The compiler first checks the static legality constraints from \sref{sec:co-compilation}.
A proposal is rejected before compilation if, for example, its conjugated operations are dependent, its joint placement introduces a cycle, a group spans more than one segment of either model, or its required operands exceed the memory budget.
Each rejection is returned to the agent through a specific code in $\mathbf{c}_t$.

For every new conjugated group that passes these checks, the compiler builds and tunes its cross-model kernel and applies a kernel-gain gate before inserting it into the round.
The candidate kernel and the separately compiled kernels of its members are replayed as captured CUDA graphs in alternating timing blocks, with weights rotated beyond the L2 cache between measurements.
A group $g$ is admitted only if
\begin{equation}
    v(g)
    =
    \frac{\sum_k \tau(o_k)}
         {\widehat{\tau}(g)}
    >
    1+\max\!\left(\delta,z\widehat{s}\right),
    \label{eq:kernel-gain}
\end{equation}
where $\widehat{\tau}(g)$ includes all work launched by the group, $\delta$ is a minimum gain margin, and $\widehat{s}$ is the standard error of
the paired timing relative to $\widehat{\tau}(g)$.
Every result, whether the group passes or fails, enters $\mathbf{v}_t$ so the agent can learn which conjugations produce useful local improvements.

Passing the gate establishes only that a conjugated group is locally promising.
The resulting joint optimization plan is therefore compiled into the complete co-scheduled round and checked against the reference execution for every draft step.
We require relative agreement within $10^{-2}$ and permit an argmax difference only when the reference logits themselves form a near tie; failing positions or joint sets are quarantined and reported to the agent.

We then measure each candidate joint optimization plan on the complete co-scheduled round, using co-scheduling without co-compilation as the anchor.
Each timing block replays the anchor, candidate, placement control, candidate, and anchor in that order.
The placement control preserves the candidate's join and fork dependencies but executes the original separately compiled kernels.
Comparing the anchor with this control measures the effect of the scheduling arrangement; comparing the control with the candidate measures the additional effect of the compiler transformations.

We compute the candidate's full-round speedup over the anchor as the median of the paired anchor-to-candidate latency ratios.
To express the search reward relative to baseline, we multiply this speedup by the anchor's speedup over the faster of two conventional baselines: the compiled models executed back-to-back, or sequentially within one captured CUDA graph.
This scaling factor is fixed within a configuration, so it does not change how MCTS ranks candidate plans.

The measured full-round reward updates the MCTS search tree, while kernel-level verdicts and compiler feedback enter the LLM's accumulated optimization experience to guide subsequent proposals.
Tuned kernels, kernel verdicts, compiled joint nodes, rejection outcomes, and search checkpoints persist across escalation phases and resumed runs.
The system therefore improves through accumulated compiler and GPU experience, without modifying the weights of the LLM agent.
Appendix~\ref{sec:appendix-search} details search-space expansion and final plan acceptance.

\section{Evaluation} \label{sec:eval}

\vspace{-10pt}
\subsection{Methodology} \label{sec:eval:meth}
\vspace{-2pt}
\niparagraph{Benchmarks.}
We evaluate four target-draft pairs on NVIDIA B300 GPU: Llama-3.1-8B/Llama-3.2-1B, Qwen3-14B/Qwen3-0.6B, Qwen3-32B/Qwen3-Instruct-2507-4B, and Llama-3.1-70B/Llama-3.1-8B~\citep{llama31, llama32, qwen3technicalreport}.
We evaluate Llama-3.2-3B/Llama-3.2-1B and Qwen3-Instruct-2507-4B/Qwen3-0.6B on NVIDIA RTX 5090 GPU.
All experiments use batch size 8~\citep{bass:acl:2024}, speculation length $\gamma\leq6$~\citep{spec-decode:icml:2023, bass:acl:2024, pearl:iclr:2025}, a 1024-token input sequence length~\citep{splitwise:isca:2024}, and FP16 precision~\citep{bass:acl:2024}.
\vspace{-2pt}
\niparagraph{Evaluated systems.}
%
\Baseline{} represents the current practice for deploying speculative-decoding models, where the draft and target models are compiled independently and executed sequentially~\citep{spec-decode:icml:2023,vllm:sosp:2023}.
\Conjugate{} co-locates the draft and target models on the same GPU and jointly optimizes cross-model compilation plans using measurements of their shared execution through the closed-loop search of \sref{sec:agentic-search}.
A dependency-preserving multi-stream schedule realizes legal concurrent execution, including between non-matching operators such as Tensor Core GEMMs and CUDA Core operations.
\ConjugateInt{} additionally replaces selected compatible operation groups with kernels that interleave their work within one launch.

\vspace{-2pt}
\niparagraph{Implementation.}
We implement \Conjugate{} in Python using Apache TVM v0.25~\citep{tvm} to import PyTorch~\citep{pytorch} models, lower their computation, and generate CUDA code.
Our search forms a closed loop between the agent, compiler, and GPU, where agent proposes cross-model optimization plans, compiler turns them into joint nodes and interleaved kernels that it tunes and builds on the TVM backend, and measurements and compiler feedback inform subsequent proposals.
We organize the evaluated plans in a Monte Carlo tree search (MCTS)~\citep{mcts}, which propagates their measured rewards through the tree to prioritize subsequent exploration.
We use OpenAI's gpt-5.6-luna~\citep{gpt56} as the agent. 
\vspace{-2pt}
\niparagraph{Measurements.}
We run all experiments with CUDA 12.8 on NVIDIA B300 SXM6 and RTX 5090 GPUs.
We run experiments 50 times, report the median, and normalize all results to \Baseline.
For GPU utilization we use Nsight Compute~\citep{nsight-compute} and collect kernel traces with CUPTI~\citep{cupti}.
\subsection{Experimental Results} \label{sec:eval:results}

\edef\evalSavedIntextsep{\the\intextsep}
\setlength{\intextsep}{4pt}
\begin{figure}[h]
  \centering
  \setlength{\abovecaptionskip}{3pt}
  \setlength{\belowcaptionskip}{0pt}
  \begin{minipage}[b]{0.55\linewidth}
    \centering
    \includegraphics[width=0.85\linewidth]{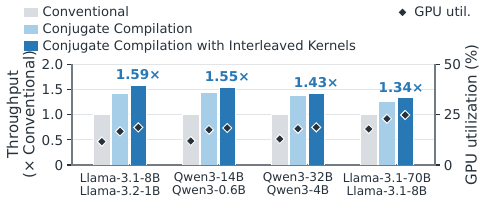}
  \end{minipage}\hfill
  \begin{minipage}[b]{0.37\linewidth}
    \centering
    \includegraphics[width=0.8\linewidth]{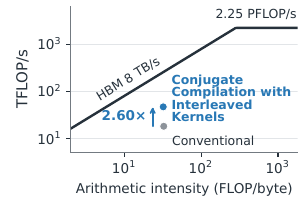}
  \end{minipage}
  \par\nointerlineskip
  \begin{minipage}[t]{0.5\linewidth}
    \caption{Throughput relative to \Baseline{} (left y-axis) and GPU utilization (right y-axis) on NVIDIA B300.}
    \label{fig:results-throughput}
  \end{minipage}\hfill
  \begin{minipage}[t]{0.37\linewidth}
    \caption{Roofline analysis of Llama down-projection kernel on B300 GPU.}
    \label{fig:roofline}
  \end{minipage}
\end{figure}

\niparagraph{Throughput improvement.} \label{sec:results-throughput}
\fref{fig:results-throughput} shows target-model verification throughput relative to \Baseline{}.
Across the four benchmarks, \Conjugate{} improves throughput by $1.38\times$ on average over \Baseline{}.
The improvement reflects cross-model compilation informed by shared-execution measurements, enabling joint optimization of the draft and target models rather than optimizing each model independently (see \sref{sec:co-compilation} and \sref{sec:agentic-search}).
\ConjugateInt{} further raises throughput to $1.47\times$ that of \Baseline{} on average, achieving the highest throughput in every benchmark and reaching $1.59\times$ on Llama-3.1-8B/Llama-3.2-1B.
The additional 7\% average gain over \Conjugate{} is due to co-optimizing selected draft and target operations within a common interleaved kernel, beyond co-scheduling separate kernels (\sref{sec:co-compilation}).
\niparagraph{GPU utilization.}
To understand the throughput improvements, we plot GPU utilization in \fref{fig:results-throughput}.
Across the four benchmarks, \ConjugateInt{} increases utilization by 50\% on average over \Baseline{}.
Consider Llama-3.1-8B/Llama-3.2-1B, where utilization increases from 11.6\% to 18.7\%.
Co-located shared execution allows ready draft operations to use GPU resources left available during target execution, rather than requiring the two models to execute sequentially.
Interleaved kernels further allow the compiler to schedule selected work from both models within a single launch, so work from one operation can fill execution capacity left unused by the other (\sref{sec:co-compilation}).
The Llama pair's 61\% utilization increase is consistent with more effective use of the GPU during joint execution and supports the 59\% increase in throughput.
%


%
\niparagraph{Roofline analysis.} \label{sec:results-roofline}
To examine how interleaving improves kernel performance, \fref{fig:roofline} compares the down projections of Llama-3.1-8B/Llama-3.2-1B under \Baseline{} and \ConjugateInt{} on B300.
The two \Baseline{} kernels achieve a combined throughput of 17.9 TFLOP/s when executed sequentially, whereas the interleaved kernel achieves 46.5 TFLOP/s, a $2.6\times$ improvement.
The interleaved kernel operates closer to the memory-bandwidth roof, indicating better effective bandwidth use when the two projections are compiled and executed together.
%


\begin{figure}[!htbp]
  \centering
  \setlength{\abovecaptionskip}{3pt}
  \setlength{\belowcaptionskip}{0pt}
  \begin{minipage}[b]{0.32\linewidth}
    \centering
    \includegraphics[width=\linewidth]{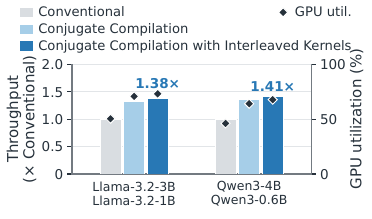}
  \end{minipage}\hfill
  \begin{minipage}[b]{0.32\linewidth}
    \centering
    \includegraphics[width=\linewidth]{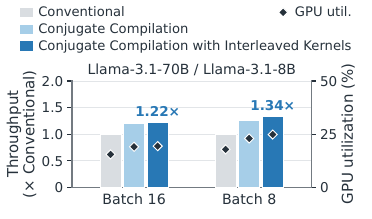}
  \end{minipage}\hfill
  \begin{minipage}[b]{0.32\linewidth}
    \centering
    \includegraphics[width=\linewidth]{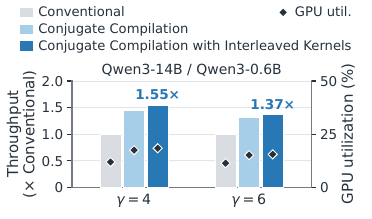}
  \end{minipage}
  \par\nointerlineskip
  \begin{minipage}[t]{0.32\linewidth}
    \caption{Sensitivity to GPU platform.}
    \label{fig:results-throughput-5090}
  \end{minipage}\hfill
  \begin{minipage}[t]{0.32\linewidth}
    \caption{Sensitivity to batch size.}
    \label{fig:results-batch-size}
  \end{minipage}\hfill
  \begin{minipage}[t]{0.32\linewidth}
    \caption{Sensitivity to draft length.}
    \label{fig:results-throughput-q14-gamma}
  \end{minipage}
\end{figure}


%
\niparagraph{Sensitivity to GPU platform.}
\fref{fig:results-throughput-5090} evaluates Llama-3.2-3B/Llama-3.2-1B and Qwen3-4B/Qwen3-0.6B on NVIDIA RTX 5090 GPU.
\Conjugate{} improves throughput by $1.33\times$ and $1.37\times$ over \Baseline{}, respectively, while \ConjugateInt{} raises the speedups to $1.38\times$ and $1.41\times$, averaging $1.40\times$ across the two pairs.
Unlike the B300 workloads, the RTX 5090 workloads already achieve substantially higher utilization under \Baseline{} (50.4\% for the Llama pair and 46.2\% for the Qwen pair).
Nevertheless, cross-model compilation informed by shared-execution measurements remains beneficial at higher baseline utilization, with \ConjugateInt{} raising utilization to 73.2\% and 67.9\%, respectively.
Co-location and execution-informed compilation provide most of the throughput improvement, while interleaving adds a further 4.1\% and 3.3\%, extending optimization from separate kernels to common cross-model kernels.

\niparagraph{Sensitivity to batch size.}
\fref{fig:results-batch-size} evaluates Llama-3.1-70B/Llama-3.1-8B at batch sizes 8 and 16 on B300 GPU.
At batch size 8, \Conjugate{} and \ConjugateInt{} improve throughput by $1.27\times$ and $1.34\times$ over \Baseline{}, respectively, while at batch size 16, the speedups are $1.20\times$ and $1.22\times$.
The larger relative gains at batch size 8 are consistent with each model supplying less work per invocation, leaving more opportunity for co-location and interleaving to improve GPU resource use by combining work across models.
At batch size 16, more work is available within each model, allowing batching itself to contribute more to throughput while joint compilation provides a further $1.22\times$ improvement over \Baseline{}.
Batching and joint compilation therefore offer complementary benefits.
\FloatBarrier

\niparagraph{Sensitivity to draft length.}
\fref{fig:results-throughput-q14-gamma} evaluates Qwen3-14B/Qwen3-0.6B at batch size 8 with draft lengths $\gamma=4$ and $\gamma=6$.
At $\gamma=4$, \Conjugate{} and \ConjugateInt{} achieve $1.45\times$ and $1.55\times$ speedups over \Baseline{}, respectively, compared with $1.32\times$ and $1.37\times$ at $\gamma=6$.
Longer drafts add sequential draft steps while the target verifies more tokens in one pass, changing the balance of work available for overlap.
The larger gains at $\gamma=4$ are consistent with a more favorable balance between drafting and verification.
At the measured acceptance rate of $\alpha=0.73$, \ConjugateInt{} generates 852 tokens/sec at $\gamma=4$ versus 713 at $\gamma=6$, favoring $\gamma=4$ for the workload.
\vspace{-3pt}
\section{Related Work}
\label{sec:related} \vspace{-10pt}
Parallel speculative decoding systems coordinate drafting and verification through runtime scheduling and placement~\citep{pearl:iclr:2025,faser:arxiv:2026}.
GPU-sharing and context-aware compilation approaches schedule concurrent computations or optimize kernels for resource constraints~\citep{orion:eurosys:2024,veltair:asplos:2022}.
LLM-guided optimization typically targets individual kernels or programs~\citep{kernelbench:icml:2025,stark:iclr:2026}.
Conjugate Compilation searches joint draft--target compilation plans using an LLM agent and MCTS, with measurements of the complete co-located shared execution guiding subsequent proposals.
Additional related work is in Appendix~\ref{sec:appendix-related}.

\vspace{-4pt}
\section{Conclusion}
\label{sec:conclusion}
\vspace{-10pt}
We introduced Conjugate Compilation, a framework for jointly optimizing co-located models using the measured performance of their shared execution to guide compilation.
Our approach combines a self-improving LLM--MCTS search over cross-model compilation plans with transformations that bring work from multiple models into common kernels, spanning GEMM and non-GEMM operations.
Across speculative-decoding workloads, our results show that cross-model Conjugate Compilation can recover GPU capacity left stranded when models are optimized independently.



\bibliography{iclr2027_conference}

@string{asplos="ASPLOS"}

@string{cc="CC"}

@string{eurosys="EuroSys"}

@string{iclr="ICLR"}

@string{icml="ICML"}

@string{ics="ICS"}

@string{isca="ISCA"}

@string{mlsys="MLSys"}

@string{neurips="NeurIPS"}

@string{osdi="OSDI"}

@string{sosp="SOSP"}

@string{arxiv="arXiv"}

@misc{nsight-compute,
  author = {{NVIDIA}},
  title = {{NVIDIA} {Nsight} {Compute}},
  howpublished = {\url{https://docs.nvidia.com/nsight-compute/}},
  note = {Version 2025.1, shipped with CUDA Toolkit 12.8; accessed September 2026},
}

@misc{cupti,
  author = {{NVIDIA}},
  title = {{CUPTI}: {CUDA} Profiling Tools Interface},
  howpublished = {\url{https://docs.nvidia.com/cupti/12.8/index.html}},
  note = {CUPTI 12.8 documentation (v2025.1.0), accessed September 2026},
}

@misc{cuda-streams,
  author = {{NVIDIA}},
  title = {{CUDA} Programming Guide: Asynchronous Execution},
  howpublished = {\url{https://docs.nvidia.com/cuda/cuda-programming-guide/02-basics/asynchronous-execution.html}},
  note = {CUDA Programming Guide v13.4.2, accessed September 2026},
}

@misc{gpt56,
  author = {{OpenAI}},
  title = {{GPT-5.6}: Frontier Intelligence that Scales with Your Ambition},
  howpublished = {\url{https://openai.com/index/gpt-5-6/}},
  year = {2026},
}

@inproceedings{bass:acl:2024,
    title = "{BASS}: Batched Attention-optimized Speculative Sampling",
    author = "Qian, Haifeng  and
      Gonugondla, Sujan Kumar  and
      Ha, Sungsoo  and
      Shang, Mingyue  and
      Gouda, Sanjay Krishna  and
      Nallapati, Ramesh  and
      Sengupta, Sudipta  and
      Ma, Xiaofei  and
      Deoras, Anoop",
    editor = "Ku, Lun-Wei  and
      Martins, Andre  and
      Srikumar, Vivek",
    booktitle = "ACL Findings",
    year = "2024",
}

@misc{llama32,
    title = {Llama 3.2: Revolutionizing edge AI and vision with open, customizable models},
    author = {Meta},
    year = {2024},
    howpublished = {\url{https://ai.meta.com/blog/llama-3-2-connect-2024-vision-edge-mobile-devices/}}
}

@inproceedings{rmsnorm:neurips:2019,
      title={Root Mean Square Layer Normalization}, 
      author={Biao Zhang and Rico Sennrich},
    booktitle = neurips,
      year={2019}, 
}

@inproceedings{kernelbench:icml:2025,
author = {Ouyang, Anne and Guo, Simon and Arora, Simran and Zhang, Alex L and Hu, William and R{\'e}, Christopher and Mirhoseini, Azalia},
title = {{KernelBench}: can {LLMs} write efficient {GPU} kernels?},
booktitle = icml,
year = {2025},
}

@inproceedings{stark:iclr:2026,
title={{STARK}: Strategic Team of Agents for Refining Kernels},
author={Juncheng Dong and Yang Yang and Tao Liu and Yang Wang and Feng Qi and Vahid Tarokh and Kaushik Rangadurai and Shuang Yang},
booktitle=iclr,
year={2026},
}

@inproceedings{elasticroom:hpdc:2024,
author = {Ma, Lixian and Chen, Haoruo and Shao, En and Wang, Leping and Chen, Quan and Tan, Guangming},
title = {{ElasticRoom}: Multi-Tenant {DNN} Inference Engine via Co-design with Resource-constrained Compilation and Strong Priority Scheduling},
booktitle = {HPDC},
year = {2024},
}

@inproceedings{veltair:asplos:2022,
author = {Liu, Zihan and Leng, Jingwen and Zhang, Zhihui and Chen, Quan and Li, Chao and Guo, Minyi},
title = {{VELTAIR}: towards high-performance multi-tenant deep learning services via adaptive compilation and scheduling},
booktitle = asplos,
year = {2022},
}

@inproceedings{conco:ics:2025,
author = {Lu, Jiamin and Sun, Jingwei and Xu, Yunlong and Sun, Peng and Sun, Guangzhong},
title = {{ConCo}: Optimizing Compilation of Concurrent Tensor Programs on Shared {GPU}},
booktitle = ics,
year = {2025},
}

@inproceedings{gpu-interference:socc:2025,
author = {Elvinger, Paul and Strati, Foteini and Jerger, Natalie Enright and Klimovic, Ana},
title = {Understanding {GPU} Resource Interference One Level Deeper},
booktitle = {SoCC},
year = {2025},
}

@inproceedings{bless:eurosys:2025,
author = {Zhang, Shulai and Chen, Quan and Cui, Weihao and Zhao, Han and Xue, Chunyu and Zheng, Zhen and Lin, Wei and Guo, Minyi},
title = {Improving {GPU} Sharing Performance through Adaptive Bubbleless Spatial-Temporal Sharing},
booktitle = eurosys,
year = {2025},
}

@ARTICLE{opara:tc:2025,
author={Chen, Aodong and Xu, Fei and Han, Li and Dong, Yuan and Chen, Li and Zhou, Zhi and Liu, Fangming},
journal={ IEEE Transactions on Computers },
title={Opara: Exploiting Operator Parallelism for Expediting {DNN} Inference on {GPUs}},
year={2025},
volume={74},
number={01},
pages={325-333},
publisher={IEEE Computer Society},
address={Los Alamitos, CA, USA},
month=jan
}

@inproceedings{nimble:neurips:2020,
      title={Nimble: Lightweight and Parallel {GPU} Task Scheduling for Deep Learning}, 
      author={Woosuk Kwon and Gyeong-In Yu and Eunji Jeong and Byung-Gon Chun},
      booktitle = neurips,
      year={2020},
}

@article{faser:arxiv:2026,
      title={{FASER}: Fine-Grained Phase Management for Speculative Decoding in Dynamic {LLM} Serving}, 
      author={Wenyan Chen and Chengzhi Lu and Yanying Lin and Dmitrii Ustiugov},
      journal = arxiv,
      year={2026}, 
}

@inproceedings{
minedraft:icml:2026,
title={{MineDraft}: A Framework for Batch Parallel Speculative Decoding},
author={Zhenwei Tang and Arun Verma and Zijian Zhou and Zhaoxuan Wu and Alok Prakash and Daniela Rus and Bryan Kian Hsiang Low},
booktitle=icml,
year={2026},
}

@inproceedings{swiftspec:asplos:2026,
author = {Zhang, Ziyi and Jiang, Ziheng and Jiang, Chengquan and Yu, Menghan and Zheng, Size and Lin, Haibin and Liu, Xin and Hoffmann, Henry},
title = {{SwiftSpec}: Disaggregated Speculative Decoding and Fused Kernels for Low-Latency {LLM} Inference},
booktitle = asplos,
year = {2026},
}

@inproceedings{
pearl:iclr:2025,
title={{PEARL}: Parallel Speculative Decoding with Adaptive Draft Length},
author={Tianyu Liu and Yun Li and Qitan Lv and Kai Liu and Jianchen Zhu and Winston Hu and Xiao Sun},
booktitle=iclr,
year={2025},
}

@inproceedings{lithos:sosp:2025,
author = {Coppock, Patrick H. and Zhang, Brian and Solomon, Eliot H. and Kypriotis, Vasilis and Yang, Leon and Sharma, Bikash and Schatzberg, Dan and Mowry, Todd C. and Skarlatos, Dimitrios},
title = {{LithOS}: An Operating System for Efficient Machine Learning on {GPUs}},
booktitle = sosp,
year = {2025},
}

@inproceedings{orion:eurosys:2024,
author = {Strati, Foteini and Ma, Xianzhe and Klimovic, Ana},
title = {Orion: Interference-aware, Fine-grained {GPU} Sharing for {ML} Applications},
booktitle = eurosys,
year = {2024},
}

@inproceedings{pope:mlsys:2023,
  title={Efficiently scaling transformer inference},
  author={Pope, Reiner and Douglas, Sholto and Chowdhery, Aakanksha and Devlin, Jacob and Bradbury, James and Heek, Jonathan and Xiao, Kefan and Agrawal, Shivani and Dean, Jeff},
    booktitle = mlsys,
  year={2023}
}

@misc{utilyze,
    author = {{Systalyze}},
    title = {Utilyze},
    year = {2026},
    howpublished = {\url{https://github.com/systalyze/utilyze}}
}

@inproceedings{spec-decode:icml:2023,
author = {Leviathan, Yaniv and Kalman, Matan and Matias, Yossi},
title = {Fast inference from transformers via speculative decoding},
booktitle = icml,
year = {2023},
}

@misc{iea:energy-and-ai:2025,
    author = {{International Energy Agency}},
    title = {Energy and {AI}},
    howpublished = {World Energy Outlook Special Report, \url{https://www.iea.org/reports/energy-and-ai}},
    year = {2025}
}

@inproceedings{splitwise:isca:2024,
    author = {Patel, Pratyush and Choukse, Esha and Zhang, Chaojie and Shah, Aashaka and Goiri, {\'I}{\~n}igo and Maleki, Saeed and Bianchini, Ricardo},
    title = {Splitwise: Efficient Generative {LLM} Inference Using Phase Splitting},
    year = {2024},
    booktitle = isca
}

@inproceedings{vllm:sosp:2023,
    author = {Kwon, Woosuk and Li, Zhuohan and Zhuang, Siyuan and Sheng, Ying and Zheng, Lianmin and Yu, Cody Hao and Gonzalez, Joseph and Zhang, Hao and Stoica, Ion},
    title = {Efficient Memory Management for Large Language Model Serving with {PagedAttention}},
    year = {2023},
    booktitle = sosp
}

@inproceedings{reasoning-compiler:neurips:2025,
    title={{REASONING} {COMPILER}: {LLM}-Guided Optimizations for Efficient Model Serving},
    author={Annabelle Sujun Tang and Christopher Priebe and Rohan Mahapatra and Lianhui Qin and Hadi Esmaeilzadeh},
    booktitle=neurips,
    year={2025},
}

@article{litecoop:arxiv:2026,
    title={{LiteCoOp}: Lightweight Multi-{LLM} Shared-Tree Reasoning for Model-Serving Compiler Optimizations},
    author={Annabelle Sujun Tang and Christopher Priebe and Lianhui Qin and Hadi Esmaeilzadeh},
    journal={arXiv preprint arXiv:2602.01935},
    year={2026},
    url={https://arxiv.org/abs/2602.01935},
}

@article{little-dragon:arxiv:2026,
    title={How to Train Your {Little Dragon}: Teaching a Small Language Model to Construct and Refine {GPU} Optimization Scripts},
    author={Annabelle Sujun Tang and Wei Tang and Songyao Jiang and Tianhui Li and Shengzhe Xu and Jiin Woo and Kaan Ozkara and Youngsuk Park and Zhen Jia and Hadi Esmaeilzadeh},
    journal={arXiv preprint arXiv:2610.20980},
    year={2026},
    url={https://arxiv.org/abs/2610.20980},
}

@article{dnn-compiler-survey:tpds:2020,
    title = {The Deep Learning Compiler: A Comprehensive Survey},
    author = {Li, Mingzhen and Liu, Yi and Liu, Xiaoyan and Sun, Qingxiao and You, Xin and Yang, Hailong and Luan, Zhongzhi and Gan, Lin and Yang, Guangwen and Qian, Depei},
    journal = {IEEE Transactions on Parallel and Distributed Systems},
    volume = {32},
    number = {3},
    pages = {708--727},
    year = {2020}
}

@inproceedings{ios:mlsys:2021,
    title = {{IOS}: Inter-Operator Scheduler for {CNN} Acceleration},
    author = {Ding, Yaoyao and Zhu, Ligeng and Jia, Zhihao and Pekhimenko, Gennady and Han, Song},
    booktitle = mlsys,
    year={2021}
}

@inproceedings{llm-compiler:cc:2025,
    author = {Cummins, Chris and Seeker, Volker and Grubisic, Dejan and Roziere, Baptiste and Gehring, Jonas and Synnaeve, Gabriel and Leather, Hugh},
    title = {{LLM Compiler}: Foundation Language Models for Compiler Optimization},
    booktitle = cc,
    year = {2025}
}

@article{llm-compiler-optimization:arxiv:2023,
    title = {Large Language Models for Compiler Optimization}, 
    author = {Chris Cummins and Volker Seeker and Dejan Grubisic and Mostafa Elhoushi and Youwei Liang and Baptiste Roziere and Jonas Gehring and Fabian Gloeckle and Kim Hazelwood and Gabriel Synnaeve and Hugh Leather},
    journal = arxiv,
    year = {2023},
}

@inproceedings{tvm,
    title = {{TVM}: An Automated End-to-End Optimizing Compiler for Deep Learning},
    author = {Chen, Tianqi and Moreau, Thierry and Jiang, Ziheng and Zheng, Lianmin and Yan, Eddie and Cowan, Meghan and Shen, Haichen and Wang, Leyuan and Hu, Yuwei and Ceze, Luis and Guestrin, Carlos and Krishnamurthy, Arvind},
    booktitle = osdi,
    year = {2018},
}

@string{eurosys=" {ACM} European Conference on Computer Systems"}

@string{cc=" International Conference on Compiler Construction"}

@string{sosp=" {ACM} Symposium on Operating Systems Principles (SOSP)"}

@string{isca="International Symposium on Computer (ISCA)"}

@string{osdi="{USENIX} Symposium on Operating Systems Design and Implementation (OSDI)"}

@string{asplos="International Conference on Architectural Support for Programming Languages and Operating Systems"}

@article{qwen3technicalreport,
      title={Qwen3 Technical Report}, 
      author={{Qwen Team}},
      year={2025},
    journal=arxiv,
}

@misc{llama31,
    title = {Introducing {Llama 3.1}: Our most capable models to date},
    author = {Meta},
    year = {2024},
    howpublished = {\url{https://ai.meta.com/blog/meta-llama-3-1/}},
}

@article{mcts,
  title={A Survey of {Monte Carlo} Tree Search Methods},
  author={Browne, Cameron B and Powley, Edward and Whitehouse, Daniel and Lucas, Simon M and Cowling, Peter I and Rohlfshagen, Philipp and Tavener, Stephen and Perez, Diego and Samothrakis, Spyridon and Colton, Simon},
  journal={IEEE Transactions on Computational Intelligence and AI in games},
  volume={4},
  number={1},
  pages={1--43},
  year={2012},
}

@INPROCEEDINGS{resnet:cvpr:2016,
  author={He, Kaiming and Zhang, Xiangyu and Ren, Shaoqing and Sun, Jian},
  booktitle={CVPR}, 
  title={Deep Residual Learning for Image Recognition}, 
  year={2016},
}

@incollection{pytorch,
title = {{PyTorch: An Imperative Style, High-Performance Deep Learning Library}},
author = {Paszke, Adam and Gross, Sam and Massa, Francisco and Lerer, Adam and Bradbury, James and Chanan, Gregory and Killeen, Trevor and Lin, Zeming and Gimelshein, Natalia and Antiga, Luca and Desmaison, Alban and Kopf, Andreas and Yang, Edward and DeVito, Zachary and Raison, Martin and Tejani, Alykhan and Chilamkurthy, Sasank and Steiner, Benoit and Fang, Lu and Bai, Junjie and Chintala, Soumith},
booktitle = {NeurIPS},
year = {2019},
}
\bibliographystyle{formatting/iclr2027_conference}

\appendix
\clearpage

\clearpage
\section{Additional Search Details}
\label{sec:appendix-search}

\subsection{Expanding the Search Space}
\label{sec:search-escalation}

A fixed search space and tuning budget can cause exploration to stall before profitable co-optimizations are found.
When measured progress plateaus, it first retunes near-miss kernels with progressively larger budgets, then widens the candidate groups and legal placements, refines the co-schedule down to individual projection stages so joint nodes can contain more precisely selected work, and finally exposes additional kernel-template choices.
The agent is informed whenever the available search space expands.

We impose an effort floor before concluding that a kernel family offers no profitable co-optimization.
Each kernel family from both models is explored through multiple groupings, and every cross-model GEMM kernel appearing in a final candidate receives at least the combined tuning budget of the kernels it replaces.
This prevents differences in compiler effort from determining the comparison.

\subsection{Final Plan Acceptance}
\label{sec:final-acceptance}

Search-time measurements guide exploration, but the reported plan is subjected to a stricter final evaluation.
After search, we re-time the strongest candidates and the best agent-proposed plans in paired ABBA blocks against the co-scheduling anchor, while re-measuring the conventional and plain two-stream baselines in the same measurement window.
A plan is accepted only if its outputs pass the numerical checks in \sref{sec:agent-feedback}, the paired $95\%$ confidence interval of its gain excludes zero, and its improvement exceeds a noise threshold.
Finally, we remove every range of joint nodes whose leave-one-out gain does not exceed the noise threshold, unless the full plan is itself significantly faster than the reduced plan.
This final pruning requires each retained range to contribute measurable value, either independently or through its interaction with the rest of the plan.

\clearpage
\section{Additional Related Work}
\label{sec:appendix-related}
\subsection{Parallel Speculative Decoding}
Prior work on parallel speculative decoding primarily treats draft--target parallelism as a runtime scheduling, placement, or resource-allocation problem rather than a compile-time optimization problem.
Most of these systems place the draft and target models on separate GPUs, which improve performance but leave both devices underutilized: PEARL overlaps drafting and verification through adaptive parallel speculative decoding~\citep{pearl:iclr:2025}, MineDraft exploits batch-level parallelism across speculative-decoding rounds~\citep{minedraft:icml:2026}, and SwiftSpec disaggregates draft and target execution across GPU resources~\citep{swiftspec:asplos:2026}.
FASER instead is a profile-driven runtime system for parallel draft--target execution on a single GPU, using offline profiling to guide how the two models share the device~\citep{faser:arxiv:2026}.
These approaches operate after compilation: they schedule, place, or partition already-compiled model computations rather than changing how the draft and target are compiled together.
Conjugate Compilation instead treats co-located execution as a compilation problem, searching cross-model transformations and evaluating each candidate using the complete score produced by the co-located shared execution.

\subsection{GPU Sharing and Context-Aware Compilation} 
GPU sharing and inter-operator scheduling systems improve utilization by concurrently executing independent operators or models on the same device~\citep{nimble:neurips:2020,ios:mlsys:2021,opara:tc:2025,orion:eurosys:2024,bless:eurosys:2025,lithos:sosp:2025}.
These systems do not target speculative decoding; rather, they address general multi-tenant or inter-operator GPU sharing, primarily deciding which already-compiled computations should execute concurrently, while context-aware compilation approaches optimize individual kernels for a predefined resource budget or interference condition~\citep{veltair:asplos:2022,elasticroom:hpdc:2024,conco:ics:2025}.
More fundamentally, these approaches rely on profiling, analytical models, or conventional optimization policies to make their scheduling and compilation decisions; they do not use an execution-driven agentic compiler that learns how to improve co-located execution from the outcomes of its own optimization attempts.
Conjugate Compilation instead targets speculative decoding directly and couples an LLM agent with Monte Carlo tree search to iteratively propose cross-model compilation plans, compile and execute them, and use the complete score from the co-located shared execution together with kernel-level verdicts and compiler feedback to guide subsequent proposals.
This self-improving loop makes the measured behavior of the co-located execution the feedback signal for compilation, allowing the search to adapt its transformations as it discovers which optimizations help---or hurt---the models when they execute together.

\subsection{LLMs For Compiler Optimization}
Recent work uses LLMs and agents to optimize individual kernels or programs~\citep{kernelbench:icml:2025,stark:iclr:2026,llm-compiler:cc:2025,llm-compiler-optimization:arxiv:2023,reasoning-compiler:neurips:2025,litecoop:arxiv:2026,little-dragon:arxiv:2026}.
Conjugate Compilation extends agentic optimization across multiple models, using an LLM with MCTS to search joint draft--target compilation plans guided by measurements from their co-located shared execution.

\end{document}